\documentclass{article}
\usepackage{spconf,amsmath,graphicx}
\usepackage{times}
\usepackage{epsfig}
\usepackage{amssymb}
\usepackage{booktabs}
\usepackage{tabularx}
\usepackage{color}
\usepackage{colortbl}
\usepackage{subfigure}
\usepackage{multirow}
\usepackage{comment}
\usepackage{adjustbox}

\usepackage[pagebackref=true,breaklinks=true,colorlinks,bookmarks=false]{hyperref}

\newcommand{\tabref}[1]{Table \ref{#1}}

\def\eg{\emph{e.g}.}
\def\ie{\emph{i.e}.}
\def\etal{\emph{et al.}}

\title{Semantic-Aware Network for Aerial-to-Ground Image Synthesis}

\name{Jinhyun Jang \quad Taeyong Song \quad Kwanghoon Sohn\textsuperscript{\rm *}\thanks{$^{*}$Corresponding author}}
\address{School of Electrical and Electronic Engineering, Yonsei University, Seoul, Korea\\
E-mail: khsohn@yonsei.ac.kr}

\begin{document}
\ninept
\maketitle

\begin{abstract}
Aerial-to-ground image synthesis is an emerging and challenging problem that aims to synthesize a ground image from an aerial image.
Due to the highly different layout and object representation between the aerial and ground images, existing approaches usually fail to transfer the components of the aerial scene into the ground scene.
In this paper, we propose a novel framework to explore the challenges by imposing enhanced structural alignment and semantic awareness.
We introduce a novel semantic-attentive feature transformation module that allows to reconstruct the complex geographic structures by aligning the aerial feature to the ground layout.
Furthermore, we propose semantic-aware loss functions by leveraging a pre-trained segmentation network.
The network is enforced to synthesize realistic objects across various classes by separately calculating losses for different classes and balancing them.
Extensive experiments including comparisons with previous methods and ablation studies show the effectiveness of the proposed framework both qualitatively and quantitatively.
The code is publicly available at \url{https://github.com/jinhyunj/SANet}.
\end{abstract}


\let\thefootnote\relax\footnote{This work was supported by Institute of Information communications Technology Planning \& Evaluation (IITP) grand funded by the Korea government(MSIT) (No.2020-0-00056, To create AI systems that act appropriately and effectively in novel situations that occur in open worlds.)}

\vspace{-10pt}
\section{Introduction}
\label{sec:intro}
Aerial-to-ground image synthesis aims to predict corresponding ground-view image at a given aerial-view image.
It has received significant attention in the computer vision community as it can be applied to various media industries, including wide-area virtual scene generation, 3D simulation, and gaming.
However, it is a very challenging task since the aerial and ground images have an extremely different viewpoints, which makes the scene layouts and object representations in the two images completely different.

Recently, there have been attempts \cite{selectiongan, ding2020cross, deng2018like, deng2019using,xfork} to solve the problem by leveraging generative adversarial networks (GANs)~\cite{goodfellow2014generative, mirza2014conditional}.
Few methods~\cite{selectiongan, ding2020cross} impose a ground semantic map as a conditional input for the ground image.
However, these methods require semantic maps at the testing phase and the synthesized images are strongly conditioned on them, as in example-guided image synthesis methods \cite{ex1,ex2}.
Deng \etal~\cite{deng2018like, deng2019using} adopted conditional GANs~\cite{mirza2014conditional} that use vector representation extracted from aerial image to produce an appropriate ground image.
Regmi and Borji~\cite{xfork} proposed two models (X-Fork and X-Seq) that jointly generate ground images and corresponding semantic maps.
Although these works have shown plausible results, they do not handle the structural difference between the viewpoints or separately consider objects in different semantic classes, resulting in limited performance in difficult scenes which contain multiple objects and complex layout.

Other methods~\cite{crossnet,regmi2019cross,geonet} focus on transformation to convert the aerial scene layout into ground perspective.
They reduce the geometric difference between two views and mitigate the structural deformation problem.
Zhai \etal~\cite{crossnet} proposed to learn a transformation matrix that turns aerial image into ground-view panorama image.
Regmi and Borji~\cite{regmi2019cross} applied homography transformation to the aerial images and use them as inputs to synthesize the ground images.
These methods adopt coarse alignments of the entire scene layout and often fail to capture detailed transformations, yielding unsatisfactory results.
Lu \etal~\cite{geonet} proposed a differentiable geo-transformation layer based on orthogonal projection and panoramic rays by using aerial semantic and depth maps.
While this method has achieved great success, it is restricted to cases where a large number of ground truth semantic and depth maps are available.

\begin{figure}[!t]
    \centering
    \includegraphics[width=0.8\linewidth]{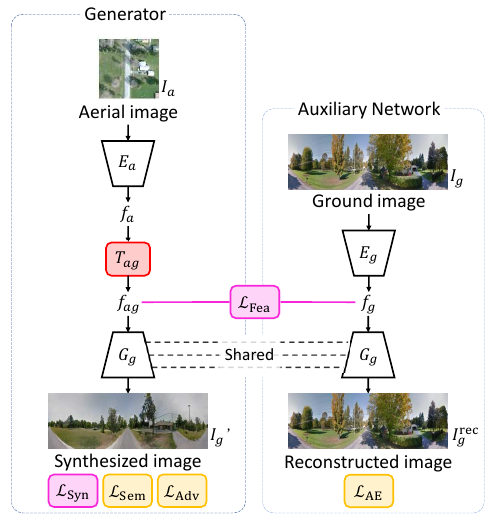}
    \vspace{-10pt}
    \caption{\textbf{Overview of our proposed framework.} Our aerial-to-ground generator is composed of an aerial encoder, a semantic-attentive feature transformation module, and a ground decoder. 
    It synthesizes a panoramic ground image given an aerial image. 
    The generator is trained using auxiliary ground autoencoder and semantic-aware loss functions.}
    \vspace{-15pt}
    \label{fig:fig1}
\end{figure}

\begin{figure*}
    \centering
    \includegraphics[width=0.95\linewidth]{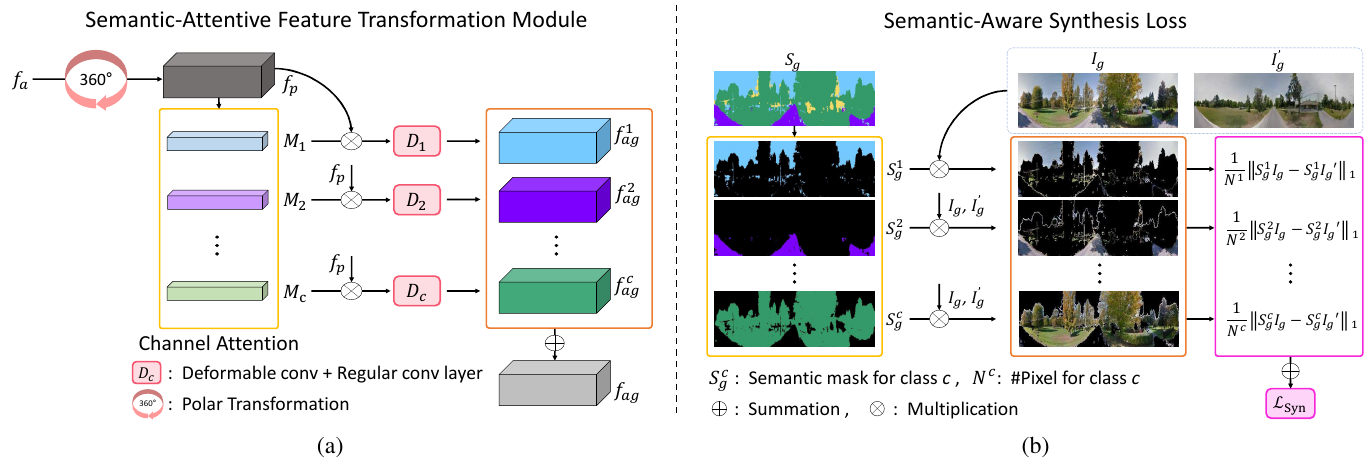}\vspace{-10pt}
    \caption{\textbf{Illustration of the proposed transformation module and semantic-aware loss function.} \textbf{(a)} Our transformation module aligns the structure of the aerial feature to the ground by applying transformation with respect to the semantic classes. \textbf{(b)} Semantic awareness of our generator is enforced by a novel loss function which balances the losses across the semantic class.
    } \vspace{-15pt}
    \label{fig:fig2}
\end{figure*}

In this paper, we propose a novel framework that imposes enhanced structural alignment with semantic awareness for aerial-to-ground image synthesis.
We argue that handling the entire scene at once is insufficient for this complex synthesis problem and therefore, explore semantically different object respectively.
To be specific, we introduce a semantic-attentive feature transformation module to align the aerial features into the ground layout.
By exploiting attention mechanism~\cite{woo2018cbam}, it separately transforms the features with different semantic representations to achieve better alignment.
Furthermore, we present semantic-aware loss functions to handle the objects across the various semantic classes.
Being aware that objects with different classes appear in uneven number of pixels and scenes, we balance the losses for different classes by leveraging a pre-trained semantic segmentation network.
Experimental results on CVUSA~\cite{cvusa} and CVACT~\cite{cvact} datasets demonstrate the effectiveness of the proposed method. \vspace{-5pt}

\section{Proposed Method}\vspace{-5pt}
\label{sec:method}
Fig.~\ref{fig:fig1} illustrates our overall framework.
Our goal is to train a deep network that synthesizes a plausible ground panorama image $I_g$ given an aerial image $I_a$.
Based on intuition that the overall layouts and semantics for different objects should be considered for synthesizing realistic images \cite{chen2019toward}, we propose a semantic-attentive feature transformation module and semantic-aware loss functions. \vspace{-8pt}

\subsection{Network Architecture} \vspace{-5pt}
\label{ssec:rel1}
Our aerial-to-ground synthesis network, \ie, generator, is composed of an aerial encoder $E_a: I_a \rightarrow f_a$ for mapping aerial image into feature space, a semantic-attentive feature transformation module $T_{ag}: f_a \rightarrow f_{ag}$ for modeling the structural changes of feature, and a decoder $G_g: f_{ag} \rightarrow I_g'$ for synthesizing the ground image.
During training, we adopt an auxiliary ground encoder $E_g: I_g \rightarrow f_g$ that maps ground image into feature space, and a pre-trained segmentation network $S_{\text{seg}}$ that extracts semantic map from ground images.
\vspace{7pt}\\
\textbf{Semantic-Attentive Feature Transformation Module.}
The proposed semantic-attentive transformation module $T_{ag}$ is illustrated in Fig.~\ref{fig:fig2}(a).
It learns a structural transformation from $f_a$ to $f_{ag}$, where $f_{ag}$ has structure aligned to $I_g$.
Rather than solely depending on an implicit learning of the transformation \cite{crossnet}, we perform initial coarse alignment using polar transformation~\cite{safa} to generate $f_p$.

Since objects with different classes are likely to locate in different areas \cite{zou2018unsupervised} (\eg, sky occupies the upper part of an image whereas road occupies the bottom), we employ semantic-attentive transformation that separately handles alignment of objects in different class.
Specifically, we generate channel attentions~\cite{woo2018cbam} $\{M_i\}_{i=1}^{c}$ for $c$ semantic classes and apply them to $f_p$.
For further alignment, we feed them into subsequent warping blocks $\{D_i\}_{i=1}^{c}$, each of which consists of a deformable convolution layer~\cite{dai2017deformable} and a convolution layer.
By summing all the attentively aligned features $\{f_{ag}^i\}_{i=1}^c$, the final semantic-attentive transformed feature $f_{ag}$ is obtained.
\vspace{7pt}\\
\textbf{Auxiliary Networks.}
The ground encoder $E_g$, along with $G_g$, operate as an autoencoder~\cite{autoencoder}, \ie, $E_g$ extracts feature $f_g$ from $I_g$ and $G_g$ reconstructs the ground image $I_g^{\text{rec}}$.
It encourages $E_g$ to extract rich ground-specific feature $f_g$ with sufficient representations to reconstruct ground images. 
We use $f_g$ as a reference for $f_{ag}$, thereby guide $E_a$ to extract representative features for synthesizing $I'_g$, as well as $T_{ag}$ to better model the structural transformations without 3D information \cite{geonet}.
We use separate parameters for $E_a$ and $E_g$ to encourage higher flexibility and capacity in feature extraction \cite{cvmnet}. 

A pre-trained semantic segmentation network $S_\text{seg}$ takes ground image $I_g$ as an input and outputs a semantic segmentation mask $S_g$.
We use it to improve the semantic-awareness of the generator through semantic-aware losses, presented in the following section. \vspace{-10pt}

\begin{figure*}[!t]
    \includegraphics[width=0.99\textwidth]{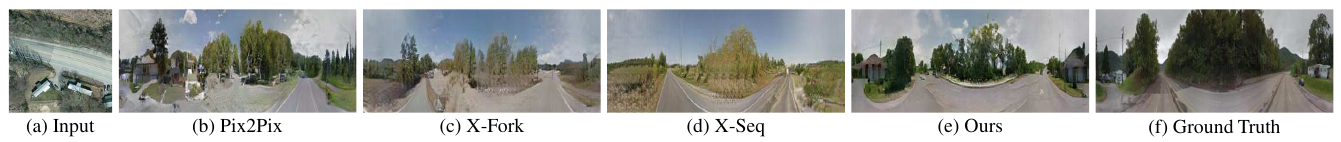}\vspace{-10pt}
    \caption{\textbf{Qualitative comparison on CVUSA dataset.}
    } \vspace{-10pt}
    \label{fig:qualitative_cvusa}
\end{figure*}

\begin{figure*}[!t]
    \includegraphics[width=0.99\textwidth]{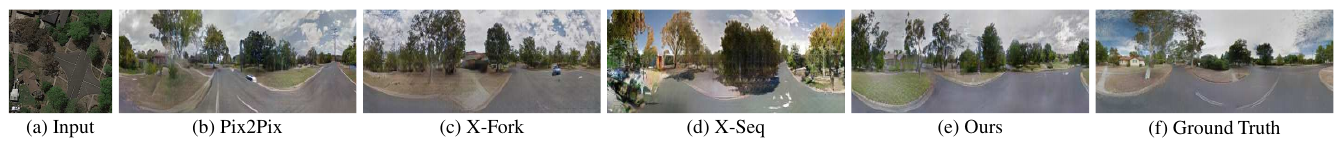}\vspace{-10pt}
    \caption{\textbf{Qualitative comparison on CVACT dataset.}
    } \vspace{-10pt}
    \label{fig:qualitative_cvact}
\end{figure*}

\subsection{Loss functions}
\label{ssec:method_loss}
\begin{figure}[!t]
    \centering
    \includegraphics[width=0.85\linewidth]{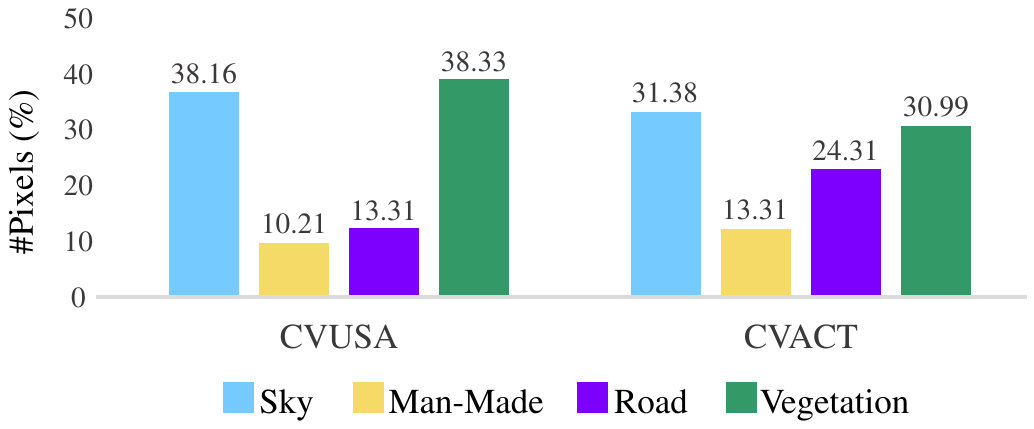}
    \vspace{-10pt}
    \caption{\textbf{Class distribution on CVUSA \cite{cvusa} and CVACT \cite{cvact} datasets.} We state the number of pixels for each class. Sky and vegetation are prevailed in the datasets whereas the man-made is few.
    }\vspace{-10pt}
    \label{fig:class_imbalance}
\end{figure}

\textbf{Semantic-Aware Synthesis Loss.}
In order to synthesize plausible ground images, objects across various classes should be considered.
Here, we observe uneven number of pixels for different objects in the cross-view image datasets \cite{cvusa, cvact} as reported in Fig. \ref{fig:class_imbalance}.
This is mainly due to the different sizes of the objects and a prevailing number of scenes without a man-made object.
It leads the regular L1 loss between the ground-truth and synthesized images to be dominated by the prevailing objects.
Consequently, the model often fails to synthesize the objects across various semantic class.

To alleviate this problem, we balance the losses across the semantic classes by exploiting $S_{g}$.
Concretely, we compute L1 loss for each class independently using the segmentation mask and average them with their own number of pixels as illustrated in Fig.~\ref{fig:fig2}(b).
Our novel semantic-aware synthesis loss is defined as
\vspace{-3pt}
\begin{equation}\label{semanticawareimg}
    \mathcal{L}_{\text{Syn}}=\sum\limits_{i=1}^c{\frac{w_i}{N_i}\left\| S_g^iI_g-S_g^iI_g' \right\|_1 },
\end{equation}
where $N_i$ and $S_g^i$ are the number of pixels and class-mask for class $i$.
We further use $w_i$ as class balancing weight~\cite{lin2017focal,ren2018learning} to handle a prevailing number of scenes without a man-made object.
\vspace{7pt}\\
\textbf{Semantic-Aware Feature Loss.}
Similar to semantic-aware synthesis loss, we use downsampled $S_g$ and apply separate loss functions for each semantic-attentive branch in the transformation module as
\vspace{-3pt}
\begin{equation}\label{semanticawarefeat}
\vspace{-3pt}
    \mathcal{L}_{\text{Fea}}=\sum\limits_{i=1}^c{\frac{w_i}{N_i}\left\| S_g^if_g-S_g^if_{ag}^i \right\|_1 },
\end{equation}
so that each channel attention is learned to highlight class-specific features and enables an effective learning of the transformation.
\vspace{7pt}\\
\textbf{Semantic Consistency Loss.}
To enforce the semantic consistency in the synthesized image, we constrain the semantic differences between the synthesized image and the ground truth image, defined as
\vspace{-3pt}
\begin{equation}\label{semanticconsistency}
    \mathcal{L}_{\text{Sem}}=\left\| S_g-S_g' \right\|_1,
\end{equation}
where $S'_g$ is output of $S_\text{seg}$ with $I'_g$ as input.
Different from the previous works \cite{selectiongan,xfork,geonet}, the parameters in $S_\text{seg}$ are fixed when training the generator.
It further encourages the synthesized objects to have similar appearance as the ground truth images $S_\text{seg}$ is trained on.
\vspace{7pt}\\
\textbf{Ground Image Autoencoding Loss.} 
In order to train $E_g$ to extract ground-representative features $f_g$, and $G_g$ to synthesize realistic ground images upon the feature, we adopt L1 reconstruction loss between $I_g$ and $I_g^{\text{rec}}$ as
\vspace{-3pt}
\begin{equation}
    \mathcal{L}_{\text{AE}}=\left\| I_g-I_g^{\text{rec}} \right\|_1.
\end{equation}
\textbf{Adversarial Loss.} 
Similar to the previous works for image synthesis \cite{pix2pix,cyclegan}, we encourage the synthesized images to be indistinguishable from the real images by adopting a discriminator $D$ and applying an adversarial loss as
\vspace{-3pt}
\begin{equation}
    \mathcal{L}_{Adv}=\mathbb{E}_{I_g}\log D(I_g)+\mathbb{E}_{I_g'}\log (1-D(I_g')).
\end{equation}

\noindent\textbf{Overall Loss.} 
In summary, our full objective is defined as
\vspace{-3pt}
\begin{equation}\label{totalloss}
\begin{aligned}
\mathcal{L}=\lambda_{\text{Syn}}\mathcal{L}_{\text{Syn}} + \lambda&_{\text{Fea}}\mathcal{L}_{\text{Fea}} + \lambda_{\text{Sem}}\mathcal{L}_{\text{Sem}}\\ + \lambda_{\text{AE}}\mathcal{L}_{\text{AE}} + \mathcal{L}&_{Adv},
\end{aligned}
\end{equation}
where $\lambda_{\text{Syn}}, \lambda_{\text{Fea}}$, $\lambda_{\text{Sem}}$ and $\lambda_{\text{AE}}$ are hyper-parameters that control the weights between the different loss terms.

\section{Experiments}
\label{sec:experiments}

\subsection{Implementation and Experimental Settings}
\textbf{Network Architecture.}
We adopt the architecture from Zhu \etal~\cite{cyclegan} for our generator and discriminator.
Specifically, we compose the encoders and decoder with 4 and 5 residual blocks~\cite{resnet} respectively.
We use SegNet~\cite{segnet} architecture for $S_{\text{seg}}$.
\vspace{7pt}\\
\textbf{Datasets.}
We conduct experiments on two commonly used cross-view image datasets, CVUSA~\cite{cvusa} and CVACT~\cite{cvact}, each containing 35,532/8,884 and 35,532/92,802 train/test image pairs.
We apply pre-processing to adjust aerial images into a similar scale and exclude ground image areas with large panoramic distortions.
Then we resize the aerial and ground images into $256\times256$ and $128\times512$.
For both datasets, we use $S_{\text{seg}}$ trained on CVUSA dataset, with the segmentation maps provided by \cite{crossnet} as pseudo label which contains four classes of sky, man-made, road and vegetation.
\vspace{7pt}\\
\textbf{Training Details.}
We set the class balancing weights $w_i$ in (\ref{semanticawareimg}) and (\ref{semanticawarefeat}) as $0.5, 2, 1,$ and $1$ for sky, man-made, road, and vegetation classes, respectively. 
We aim to handle the scarcity of the man-made class and avoid the network being overfitted to the ground-truth sky representation.
We set the loss weights in (\ref{totalloss}) as $\lambda_{\text{Syn}}=10, \lambda_{\text{Fea}}=2, \lambda_{\text{Sem}}=2$, and $\lambda_{\text{AE}}=5$.
We use Adam optimizer \cite{adam} with momentum parameters 0.5 and 0.999, and fixed learning rate of 0.0002.
The network parameters are initialized with normal distribution with zero mean and 0.02 standard deviation.
We do not perform any data augmentation and train our network for 30 epochs with batch size of 4.
All the experiments are conducted using Pytorch \cite{pytorch} library, on a single NVIDIA RTX 2080Ti X GPU.
\vspace{7pt}\\
\textbf{Evaluation Protocols.}
For quantitative evaluations, we follow the protocols presented in~\cite{xfork}.
We measure the visual quality of the synthesized images by Peak-Signal-to-Noise Ratio (PSNR), Structural-Similarity Index (SSIM), and Sharpness Difference (SD) with ground-truth image.
We also measure the realism and diversity of the synthesized images by Inception Score (IS), Top-k prediction accuracy, and KL divergence.
We additionally evaluate the pixel-wise semantic consistency of the synthesized images using mean Intersection-over-Union (mIoU). \vspace{-10pt}

\begin{table*}[!t]
		\caption{Quantitative evaluation of PSNR, SSIM, Sharpness Difference, KL Loss and mIoU.}
		\vspace{4pt}
		\begin{center} \label{tab:quantitative1}
			\begin{tabular}{ccccccccccc}
				\toprule
				\multirow{2}{*}{Methods} & \multicolumn{5}{c}{CVUSA} & \multicolumn{5}{c}{CVACT} \\
				\cmidrule(lr){2-6}\cmidrule(lr){7-11}
				& PSNR & SSIM & SD & KL & mIoU & PSNR & SSIM & SD  & KL & mIoU\\
				\midrule \midrule
				Pix2Pix \cite{pix2pix}& 19.0631 & 0.3864 & 17.8758 & 4.64$\pm$1.18 & 0.3013 & 19.5376 & 0.4022 & 17.4920 & 3.64$\pm$0.93 & 0.3048\\
				X-Fork \cite{xfork}& \textbf{19.7425} & 0.4106 & 18.1640 & 4.91$\pm$1.24 & 0.2962 & \textbf{20.1629} & 0.4134 & 17.7542 & 3.55$\pm$0.90 & 0.3005\\
				X-Seq \cite{xfork}& 19.6859 & 0.4292 & 18.2379 & 6.42$\pm$1.38 & 0.2944 & 18.8307 & 0.4062 & 17.6511 & 4.13$\pm$1.03 & 0.2798\\
			    Ours & 19.6604 & \textbf{0.4363} & \textbf{18.2497} & \textbf{3.66$\pm$1.04} & \textbf{0.3068} & 19.6944 & \textbf{0.4168} & \textbf{17.9001} & \textbf{3.44$\pm$0.93} & \textbf{0.3118}\\
				\bottomrule 
			\end{tabular}\vspace{-20pt}
		\end{center}
	\end{table*}

\begin{table*}[!t]
		\caption{Quantitative evaluation of inception score and classification accuracy.}
		\vspace{4pt}
		\begin{center} \label{tab:quantitative2}
			\begin{tabular}{cccccccccccc}
				\toprule
				\multirow{4}{*}{Methods} & \multicolumn{5}{c}{CVUSA} & \multicolumn{5}{c}{CVACT} \\
				\cmidrule(lr){2-6}\cmidrule(lr){7-11}
				 & \multicolumn{3}{c}{Inception score} & \multicolumn{2}{c}{Accuracy}& \multicolumn{3}{c}{Inception score} & \multicolumn{2}{c}{Accuracy} \\
				 \cmidrule(lr){2-4} \cmidrule(lr){5-6} \cmidrule(lr){7-9} \cmidrule(lr){10-11}
				& All & Top-1 & Top-5 & Top-1 & Top-5 & All & Top-1 & Top-5 & Top-1 & Top-5 \\
				\midrule \midrule
				Pix2Pix \cite{pix2pix}& 2.2454 & 2.0252 & 2.2045 & 29.43 & 67.66 & 1.7930 & 1.6808 & 1.8094 & 23.48 & 65.05\\
				X-Fork \cite{xfork}& 2.4556 & 2.1217 & 2.4857 & 29.82 & 69.99 & 1.9412 & 1.7042 & 1.9686 & 25.41 & \textbf{67.03}\\
				X-Seq \cite{xfork}& 2.2055 & 2.0558 & 2.1902 & 24.83 & 63.70 & 2.1648 & 1.7772 & 2.1115 & 19.88 & 57.39\\
			    Ours & \textbf{2.5367} & \textbf{2.1429} & \textbf{2.5087} & \textbf{34.48} & \textbf{72.58} & \textbf{2.1762} & \textbf{1.8577} & \textbf{2.1293} & \textbf{26.24} & 63.78\\
			    \midrule
			    Real Data & 3.2930 & 2.5634 & 3.2235 & - & - & 2.4226 & 2.0046 & 2.4087 &- & -\\
				\bottomrule 
			\end{tabular}\vspace{-15pt}
		\end{center}
\end{table*}

\begin{figure*}[!t]
    \includegraphics[width=0.99\textwidth]{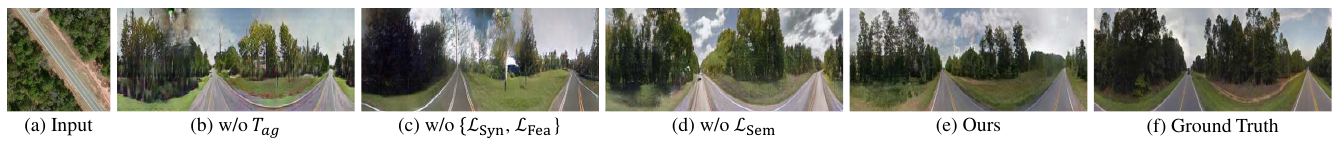}\vspace{-10pt}
    \caption{\textbf{Qualitative ablation study results on CVUSA dataset.}
    } \vspace{-15pt}
    \label{fig:ablation}
\end{figure*}

\begin{table}[]
	\caption{Quantitative ablation study results on CVUSA dataset.}\vspace{4pt}
	\begin{center} \label{tab:ablation}
			\begin{tabular}{lcccccc}
				\toprule
				\multirow{2}{*}{Setup} & \multicolumn{4}{c}{CVUSA} \\
				\cmidrule(lr){2-5}
				& SSIM & KL &  IS (All) & mIoU\\
				\midrule \midrule
				w/o $T_{ag}$& 0.4246 & 5.47$\pm$1.21 & 2.4838 & 0.2823\\
				w/o \{$\mathcal{L}_{\text{Syn}}, \mathcal{L}_{\text{Fea}}$\}& 0.3884 & 5.77$\pm$1.29 & 2.4989 & 0.2790\\
				w/o $\mathcal{L}_{\text{Sem}}$& 0.4217 & 3.85$\pm$1.29 & \textbf{2.5501} & 0.2939\\ 
				Ours& \textbf{0.4363} & \textbf{3.66$\pm$1.04} & 2.5367 & \textbf{0.3068}\\
				\bottomrule 
			\end{tabular}
	\end{center}\vspace{-15pt} 
\end{table}

\subsection{Comparison with State-of-the-Art Methods}
We compare the proposed method with Pix2Pix~\cite{pix2pix}, X-Fork~\cite{xfork}, and X-Seq~\cite{xfork}.
Since these methods handle input and output images of same shapes, we apply few modifications.
We change the bottleneck kernel size from (4, 4) to (1, 4) and use unconditional discriminator.
We also remove the skip connections in Pix2Pix~\cite{pix2pix}.
Except the above modifications, we follow the original settings.

We present qualitative results in Figs. \ref{fig:qualitative_cvusa} and \ref{fig:qualitative_cvact}. 
Compared to the previous methods \cite{pix2pix,xfork}, we observe that our method shows the most visually plausible results.
Specifically, our results show the clearest image appearance and consistent layout with the ground-truth images.
It demonstrates that our semantic-attentive feature transformation module successfully aligns the intermediate features to the ground layout by focusing on every different semantic class.
We also observe that the proposed method synthesizes plausible result across various objects, while others failed (\eg, buildings). 
This confirms that our semantic-aware loss functions allow the network to handle objects across various classes.

Quantitative results are presented in Tables \ref{tab:quantitative1} and \ref{tab:quantitative2}.
The proposed method outperforms the previous methods in all the quantitative measures except for PSNR.
Although our PSNR results are slightly lower compared to X-Fork \cite{xfork}, we achieve higher visual quality scores (KL and IS), showing that our method generates more realistic images.
In addition, we can see that our method results in the highest mIoU score which verifies that our framework generates the most semantically consistent images with the ground truth.
\vspace{-5pt}
\subsection{Ablation Study}
To investigate the importance of the key components in our framework, we conduct experiments on our method without $T_{ag}$, without \{$\mathcal{L}_{\text{Syn}}$, $\mathcal{L}_{\text{Fea}}$\}, and without $\mathcal{L}_{\text{Sem}}$ on CVUSA dataset.
Results are presented in \tabref{tab:ablation} and Fig. \ref{fig:ablation}.

For the setup w/o $T_{ag}$, we only apply polar transformation to $f_a$ and do not use any loss function for the transformed feature.
For the setup w/o \{$\mathcal{L}_{\text{Syn}}$, $\mathcal{L}_{\text{Fea}}$\}, we use regular L1 loss for images and features.
We observe that the networks with those setups generate the synthesized images that have unclear structure alignment and are not realistic, compared to our full framework.
This observation coincides with the quantitative results, indicating that both components largely affect the results and should be applied cooperatively.
For the setup w/o $\mathcal{L}_{\text{Sem}}$, both the qualitative and quantitative results show minor difference from our full framework.
It demonstrates that our transformation module, together with semantic-aware loss functions sufficiently align the structure and enhance semantic awareness to the network, thereby enforcing the semantic consistency in the synthesized image.

\section{Conclusion}
\label{sec:conclusion}
In this paper, we proposed a novel framework for aerial-to-ground image synthesis through enforcing the semantic awareness of the network.
The proposed semantic-aware network contains a novel semantic-attentive feature transformation module and is trained with semantic-aware loss functions.
The transformation module is modeled to align the structures of every object into the corresponding ground layout.
Semantic awareness is further enhanced by the proposed loss functions designed to independently calculate losses across every semantic.
By handling every semantic categories respectively, our approach succeeded in synthesizing plausible ground image from given aerial image.
Extensive experimental results demonstrate the effectiveness of the proposed method in terms of realism and semantic consistency of the synthesized images.
\let\thefootnote\relax\footnote{This research was supported by the Yonsei University Research Fund of 2021 (2021-22-0001).}

\newpage

\bibliographystyle{IEEEbib}
\bibliography{strings,refs}

\end{document}